\documentclass[11pt]{article}%
\usepackage{amsfonts}
\usepackage{amssymb}
\usepackage{graphicx}
\usepackage{setspace}
\usepackage[numbers]{natbib}
\usepackage{amsmath}%
\usepackage{amsthm}%
\usepackage{palatino}
\usepackage{paralist}
\usepackage{subfigure}
\usepackage{multirow}
\usepackage{booktabs}
\usepackage{dsfont}
\usepackage{indentfirst}
\usepackage{enumerate}
\usepackage{longtable}
\usepackage{caption}
\usepackage[hyperindex,breaklinks]{hyperref}
\hypersetup{colorlinks=true,       % false: boxed links; true: colored links
   linkcolor=red,       
    citecolor=blue,        % color of links to bibliography
    filecolor=magenta,      % color of file links
    urlcolor=cyan           % color of external links
}  

\usepackage{setspace}
\title{Sparse Regularization in Marketing  and Economics\thanks{We would like to thank Miaoyu Yang for sharing the data.}}
\author{
	\begin{tabular}{*{4}{c}p{.25\paperwidth}}
	{\large Guanhao Feng\thanks{%
			Address: 83 Tat Chee Avenue, Kowloon Tong, Hong Kong. E-mail address: 
			\texttt{gavin.feng@cityu.edu.hk}.}} & {\large  Nicholas G. Polson \thanks{
				Address: 5807 S Woodlawn Avenue, Chicago, IL 60637, USA. E-mail address: 
				\texttt{ngp@chicagobooth.edu}.}} & {\large  Yuexi Wang \thanks{
				Address: 5747 S Ellis Avenue, Chicago, IL 60637, USA. E-mail address: 
				\texttt{yxwang99@uchicago.edu}.}} & { \large Jianeng Xu \thanks{
				Address: 5807 S Woodlawn Avenue, Chicago, IL 60637, USA. E-mail address: 
				\texttt{jianeng@uchicago.edu}.}} \tabularnewline
	{\footnotesize College of Business} & 	{\footnotesize Booth School of Business} &{\footnotesize  Department of Statistics} &	{\footnotesize Booth School of Business}\tabularnewline
	{\footnotesize City University of Hong Kong} &{\footnotesize University of Chicago}& {\footnotesize University of Chicago}&{\footnotesize  University of Chicago}
	\end{tabular}
}

\date{\today}

\begin{document}
	
\maketitle
%%%%%%%%%%%%%%%%%%%%%
\begin{abstract}
%%%%%%%%%%%%%%%%%%%%%
\noindent Sparse alpha-norm regularization has many data-rich applications in Marketing and Economics. Alpha-norm, in contrast to lasso and ridge regularization, jumps to a sparse solution. This feature is attractive for ultra high-dimensional problems that occur in demand estimation and forecasting. The alpha-norm objective is nonconvex and requires coordinate descent and proximal operators to find the sparse solution. We study a typical marketing demand forecasting problem, grocery store sales for salty snacks, that has many dummy variables as controls. The key predictors of demand include price, equivalized volume, promotion, flavor, scent, and brand effects. By comparing with many commonly used machine learning methods, alpha-norm regularization achieves its goal of providing accurate out-of-sample estimates for the promotion lift effects. Finally, we conclude with directions for future research.

\end{abstract}

\newpage
\section{Introduction}
High-dimensional sparse regression is central to prediction with massive datasets and many predictors. Amazon, Facebook, Google, and Walmart employ data analytics to analyze customer behavior and to maximize customer value given purchase histories. For example, Amazon has  an "anticipatory shipping" patent to pre-ship products to reduce waiting times based on purchase order history, search history, and shopping cart activities\footnote{https://www.marketingweek.com/2014/01/22/amazon-has-seen-the-future-of-predictability/}. 
Netflix and Spotify rely heavily on extensive user databases to provide recommendations, make bids for rights of television dramas, and suggest movies to license\footnote{https://blog.kissmetrics.com/how-Netflix-uses-analytics/}. 
Uber calculates fares automatically and dynamically using GPS and street data, the rider's history, traditional routes, and other factors. Its carpooling service helps reduce average cumulative trip miles by 40\%, and the process takes millions of cars off the roads. 

Alpha-norm ($\ell_\alpha$-norm) regularization provides an attractive high-dimensional predictor selection method to handle massive datasets. Proximal algorithms can find sparse solutions as they are scalable and use Majorization-Minimization (MM) and Iterative Shrinkage-Thresholding (IST) for convergence. Our algorithm design uses coordinate-descent (\cite*{marjanovic2013exact,marjanovic2014sparsity}) together with a closed-form proximal operator of $\ell_\alpha$ penalty (\cite*{polson2015proximal}). Convergence results are provided in \cite*{attouch2013convergence} which describes the $\ell_\alpha$ objective and derives its necessary Kurdyka-{\L}ojasiewicz (KL) condition.  \cite*{zeng2014cyclic} improves this by eliminating the requirement that the columns of the design matrix be normalized, and introduces a step-size parameter to enhance the algorithm. \cite*{bredies2015minimization} provide convergence results for MM-based algorithms.

Our approach uses the closed-form proximal operator for $\ell_\alpha$ sparsity, where $0<\alpha<1$. Under $\ell_0$ sparsity, which penalizes the number of non-zero coefficients directly, Single Best Replacement (SBR) provides a fast, scalable alternative to direct posterior sampling using spike-and-slab priors, see \cite*{polson2017bayesian}. \cite*{mazumder2012sparsenet} discuss the limitations of the lasso $\ell_1$-norm versus $\ell_\alpha$-norm. The key property of $\ell_\alpha$ regularization is that it  ``jumps" to a sparse solution, which we exploit in our application.

Sparsity in marketing and economics arises due to large internet-related economic transactional data where machine-learning tools are required for estimation and forecasting. Market-demand forecasts are necessary for predicting future sales, inventory planning, and understanding the effects of potential marketing strategies. Two approaches are used to avoid in-sample overfitting: imposing an informative structure and regularization and model selection through penalizations on parameter proliferation. Model selection is especially useful with well-defined subpopulations or segments. Nonparametric approaches are used to estimate incremental effects, see \cite*{varian1982nonparametric}. Quantitative variables of interests include lift of non-targeted promotions (see \cite*{briesch2010nonparametric}), targeted promotions (see \cite*{hartmann2011identifying}), and joint lift of price and promotions predictions (see \cite*{fong2010private} and \cite*{mullainathan2017machine}).

Scanner-panel data typically includes many discrete categorical variables, such as product and store attributes, over a long time span and dummy variables to control for the individual effects. \cite*{andrews2011comparison} show that a homogeneous demand model with random in-store heterogeneity can achieve similar accuracy to model with various heterogeneity specification. Tailored methods have been developed to deal with discrete quantities and quantity discounts for packaged goods (see \cite*{allenby2004choice})  and endogenous dummy regressors, see \cite*{angrist2001estimation}.  In our empirical study, we find that $\ell_\alpha$ regularization provides better out-of-sample performance compared to linear and traditional shrinkage methods.

The rest of the paper is outlined as follows. Section \ref{alpha-regularization} shows the performance of the alpha norm in simulation studies. We demonstrate how the $\ell_\alpha$ can achieve better prediction and reduce bias in coefficient estimation in a general setting. Section \ref{simulation-study} shows the alpha norm can also produce more accurate prediction (smaller RMSE) in a simulated log-linear model and demand-estimation context.  Section \ref{empirical-analysis} applies our methodology to scanner-panel data of sales for salty snacks within one grocery store chain. Our model can achieve smaller out-of-sample RMSE and substantially shrink the number of predictors. Section \ref{discussion} concludes with directions for future research.

%%%%%%%%%%%%%%%%%%%%%%%%%%%%%%%%%%%%%%%%%
\section{Sparse $\ell_\alpha$ Regularization }\label{alpha-regularization}

Consider an ultra high-dimensional regression model for an output, $y $, with many predictors $X=(x_1, x_2, \ldots, x_p) \in \mathds{R}^{N\times p}$. The columns of $X$ are standardized such that $\Vert x_i\Vert_2 = 1$ for all $i \in \{1, 2, ... , p\}$. Hence, our model assumes:
\begin{equation}
y=X\beta+\epsilon,
\end{equation}
where $y\in \mathds{R}^{N\times 1}$ are the response observations, $\beta \in \mathds{R}^{p\times 1}$ is the sparse coefficient of interest, $X\in \mathds{R}^{N\times p}$ is the design matrix, and $\epsilon \sim N(0,\sigma^2I_N)$ is the noise. 

 The $\ell_\alpha$ regularization is equivalent  to an optimization problem which minimizes the following objective function with penalty parameter $\lambda>0$, 
\begin{equation} \label{Jbeta}
J(\beta):=\frac{1}{2}||y-X\beta||_2^2+\lambda||\beta||_\alpha^\alpha, \text{ where }  ||\beta||_\alpha:=\left(\sum_{i=1}^p|\beta_i^\alpha\right)^{\frac{1}{\alpha}}.
\end{equation}
For a given $\lambda$, we calculate a regularization path for $\hat \beta_\lambda$.

%First, we provide the proximal operator for $\ell_\alpha$-norm estimation. Consider the following scalar optimization: 
%\begin{equation}
%\underset{\beta}{\text{min}}\quad J(\beta) := \frac{1}{2}(z-\beta)^2+\lambda |\beta|^\alpha. 
%\end{equation}
Define the proximal operator, $\tau_\lambda(z)$, by $\tau_\lambda(z) = \text{argmin}_\beta \frac{1}{2}(z-\beta)^2+\lambda |\beta|^\alpha$. 
Then all its solutions are given by
\begin{equation*}
\tau_\lambda(z) =
\begin{cases}            
0, &\text{if }  |z|<h_{\lambda,\alpha}\\
\{0,\text{sgn}(z)b_{\lambda,\alpha}\}, &\text{if }  |z|=h_{\lambda,\alpha}\\
\text{sgn}(z)\bar\beta, &\text{if }  |z|>h_{\lambda,\alpha}
\end{cases}
\end{equation*}
where $b_{\lambda,\alpha}:=[2\lambda(1-\alpha)]^{\frac{1}{2-\alpha}}$ and $h_{\lambda,\alpha}:=b_{\lambda,\alpha}+\lambda \alpha b_{\lambda,\alpha}^{\alpha-1}$, and $\bar\beta>0$ satisfies $\bar\beta+\lambda \alpha\bar\beta^{\alpha-1} = |z|$. We derive two solutions, and set $\bar\beta\in(b_{\lambda,\alpha},|z|) $ to be the larger one in implementation.

\begin{figure}[!ht]
\caption{$\tau_\lambda(z)$ Function}
\begin{center}
\includegraphics[width=0.7\paperwidth]{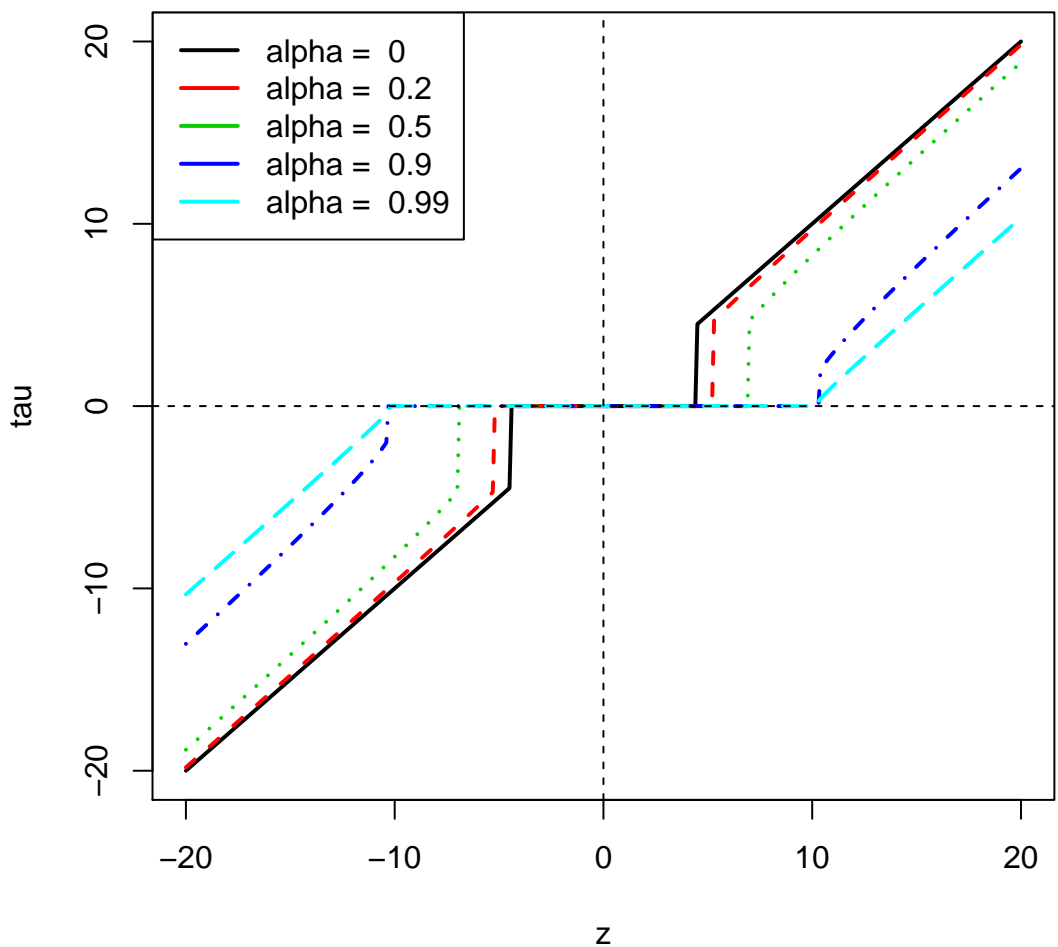}
\end{center}
\label{tau}
\smallskip
\footnotesize
Note: We show $\tau_\lambda(z)$ when $\lambda=10$ and $\alpha = 0, 0.2, 0.5, 0.9, 0.99$. $\tau_\lambda(z) = \{0, \text{sgn}(b_{\lambda,\alpha})\}$ at the point $z = h_{\lambda,\alpha}$. $\tau_\lambda(z)$ function maps the gradient $z$ to 0 when it's smaller than the threshold $h_{\lambda,\alpha}$. It jumps immediately from 0 to $b_{\lambda,\alpha}$ when $z$ arrives $h_{\lambda,\alpha}$, which results in discontinuity of coefficient regularization path.
\end{figure}

Our algorithm is based on the result of \cite*{marjanovic2012optimization} who provide a solution to $\ell_\alpha$ minimization.  First, as $\alpha\rightarrow 0^+$, the limit of $\tau_\lambda(z)$ is hard thresholding with the threshold value $h_{\lambda,0} = \sqrt{2\lambda}$, whereas as $\alpha\rightarrow 1^-$, the limit of $\tau_\lambda(z)$ is soft thresholding with the threshold value $\lambda$. For comparison, $\tau_\lambda(z)$ for different $\alpha$ are plotted in Figure \ref{tau}. The value of $\tau_\lambda(z)$ at the point $z = h_{\lambda,\alpha}$ is not unique with solution set $\{0, \text{sgn}(b_{\lambda,\alpha})\}$.

To show an improvement in the objective function,  we use Lemma 1 of \cite*{marjanovic2014sparsity}: 
\begin{equation}
\label{descent}
J(\beta_{-i}+\tau_\lambda(z_i)e_i)\leq J(\beta), \quad \text{for any } \beta.
\end{equation}
Define $z_i = z(\beta_{-i})$ as $x_i^T(y-X\beta_{-i})$ is the adjusted gradient for the $i$-th coordinate and $\beta_{-i} := \beta - \beta_ie_i$ where $e_i$ has a 1 in the $i$-th position and 0's in the rest.  Similar to the soft-thresholding rule of lasso, $\tau_\lambda(z)$ function maps the gradient $z$ to 0 when it's smaller than the threshold $h_{\lambda,\alpha}$. Therefore, $\ell_\alpha$ regularization selects a null model (all coefficient estimates are 0) if max$|(x_i,y)|< h_{\lambda,\alpha}$ (we assume $x$ and $y$ are centered and $\Vert x \Vert_2=1$).

Coordinate-descent  can then be used to iteratively minimize the objective function.  In each iteration, the algorithm minimizes along one coordinate and solves the scalar optimization problem with $z_i$ replaced by the adjusted gradient. If $\alpha=1$, this  is  equivalent to the R package {\tt glmnet} of \cite*{friedman2010regularization} .  At each step of the coordinate-descent, we use the closed-form proximal operator for $\ell_\alpha$.

%From a Bayesian perspective, the term $|\beta_j|^\alpha$ is the log-prior density for $\beta_j$. An important feature of this prior is that it concentrates more mass in the coordinate directions (\cite*{friedman2001elements}), which induces  sparsity for the posterior mode estimator.

\begin{figure}[!ht]
\caption{$\tau_\lambda(z)$ vs. $\lambda$ and $\log(h_{\lambda,\alpha})$ vs. $\log(\lambda)$}
\begin{center}
\includegraphics[width=0.4\paperwidth]{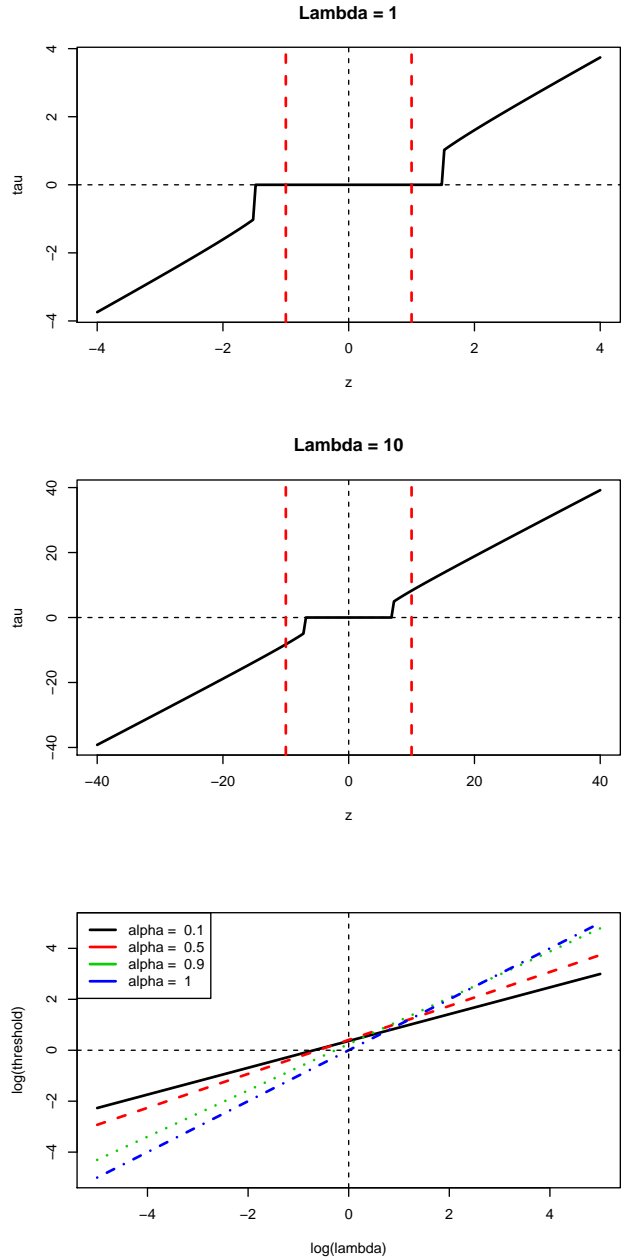}
\end{center}
\label{tau2}
\smallskip
\footnotesize
Note: The top panel is $\tau_\lambda(z)$:$\lambda=1, \alpha=0.5$.(red dashed lines are $ z = \pm \lambda$); The middle panel is $\tau_\lambda(z)$: $\lambda=10, \alpha=0.5$; The bottom panel is threshold of $\tau$ function versus $\lambda$. When $\lambda$ is small, $h_{\lambda,\alpha} > \lambda$ and when $\lambda$ is large, $h_{\lambda,\alpha} < \lambda$. The larger $\alpha$ is, the quicker $\log(h_{\lambda,\alpha})$ changes with $\log(\lambda)$. Thus estimates given by $\ell_\alpha (\alpha<1)$ are sparser than those given by lasso when we choose a small $\lambda$. On the other hand, they are more robust for too large $\lambda$'s.
\end{figure}

Unlike soft-thresholding, $\tau_\lambda(z)$ jumps immediately from 0 to $b_{\lambda,\alpha}$ when $z$ arrives to $h_{\lambda,\alpha}$, which results in discontinuity of the coefficient-regularization path. When $\lambda$ is small, $h_{\lambda,\alpha} > \lambda$, and when $\lambda$ is large, $h_{\lambda,\alpha} < \lambda$, thus the estimates are sparser than those given by lasso when we choose a small $\lambda$ (increase the shrinkage of $\hat\beta$ when true $\beta=0$). On the other hand, they are more robust for $\lambda$'s that are too large (reduce the  shrinkage of $\hat\beta$ when true $\beta\neq 0$). In Figure \ref{tau2}, we show $\tau_\lambda(z)$ when $\alpha=0.5$ and $\lambda = 1,10$. The larger $\alpha$ is, the quicker $h_{\lambda,\alpha}$ changes with $\lambda$. 

Appendix \ref{appendix} provides full details of the algorithm.

\section{Applications}\label{simulation-study}

\subsection{Linear Regression Simulation }\label{linear-model-benchmark}
To illustrate the $\ell_\alpha$-norm estimator, we simulate data from the model:

 \[y = X\beta + \epsilon, \text{ where } \epsilon \sim N(0, \sigma^2I_N), \beta \in \mathds{R}^p, \sigma^2=1,\] 
 and we minimize the objective function over a regularization path $\lambda>0$: 
\begin{equation}
J(\beta)=\Vert y-X\beta\Vert_2^2+\lambda \Vert\beta\Vert_\alpha^\alpha.
\end{equation}
The design matrices $X = [X_1, X_2, ..., X_p]$ are drawn from a multivariate normal distribution where each $X_i$ has variance $10^2$ and mean 0. We also introduce correlation among $X_i$'s, Cor$(X_i, X_j) = 0.1^{|i-j|/3}$. The noise variance $\sigma^2=1$ and $\beta$ is the coefficient vector. For three different values of $\alpha\in (0.1, 0.5, 0.9)$, we show how coefficient estimates given by $\ell_\alpha$ change with regularization parameter $\lambda$. We set three different data dimensions---$p=50, 100, \mbox{ and } 500$---whereas the true model dimension is fixed as small as 5. Specifically, $\beta = [5,5,..,5,0,0,...0]$ so that only the first five coefficients are non-zero. The number of observations is $N = 600$. 

\begin{figure}[!ht]
\caption{Simulation Example: Linear Regression Model}
\begin{center}
\includegraphics[width=0.8\paperwidth]{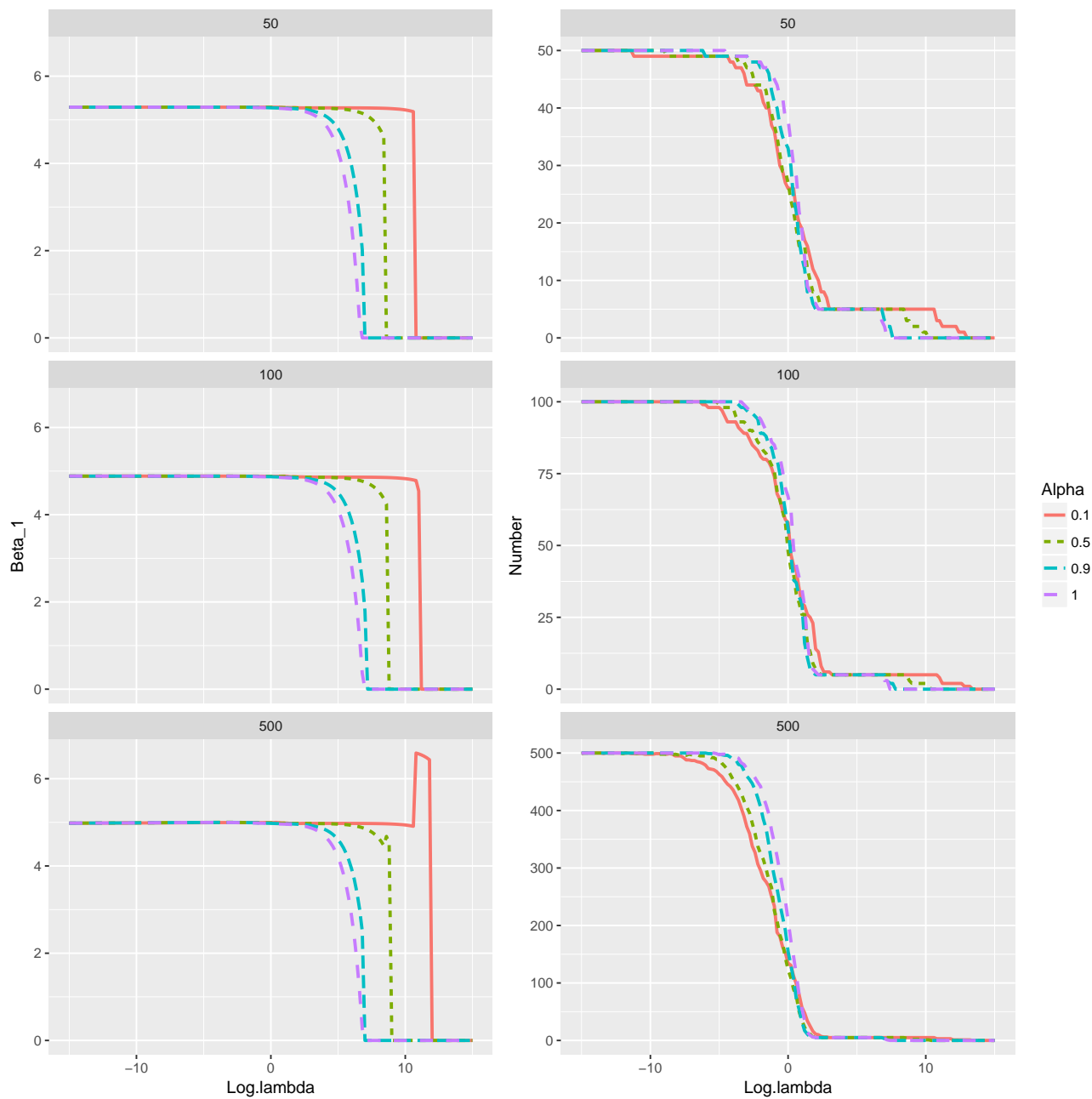}
\end{center}
\label{proxiS1}
\smallskip
\footnotesize
Note: The true number of non-zero coefficients is 5. The dataset dimension $p$ are 50, 100, 500 from top to bottom. The left column is $\hat\beta_1$ when $\beta_1=5$; The right column is the number of non-zero $\hat\beta_i$. In the left column, as $\lambda$ increases, the estimate $\hat\beta_1$ is penalized from 5 to 0. The path of lasso $(\alpha = 1)$ looks smoother and shrinks most quickly. The behavior of $\ell_{0.9}$ regularization is similar to lasso. The larger $\alpha$ is, the quicker $\hat\beta_1$ shrinks. In the right column, the path of $\ell_{0.1}$ drops most quickly when $\log(\lambda)$ is less than 0.
\end{figure}

Figure \ref{proxiS1} shows simulation results. The dataset dimension $p$'s are 50, 100, and 200 from top to bottom. The left column illustrates the case of non-zero coefficient $\beta_1 = 5$.  As $\lambda$ increases, the estimate $\hat\beta_1$ is penalized from 5 to 0. Also note the jumps in the $\ell_\alpha$-paths, especially when $\alpha=0.1$, which is expected and due to the discontinuity nature of the $\tau$ function, as discussed previously.  As the columns $X_1, X_2, ..., X_5$ are positively correlated, the drop of one $\hat\beta$ increases the estimated value of others. Figure \ref{proxiS1} shows the path of lasso ($\alpha = 1$) looks smoother and shrinks quickly. The behavior of $\ell_{0.9}$ regularization is similar to the lasso. 

\begin{figure}[!ht]
\caption{Regularization Paths for Nonzero Coefficients}
\begin{center}
\includegraphics[width=0.7\paperwidth]{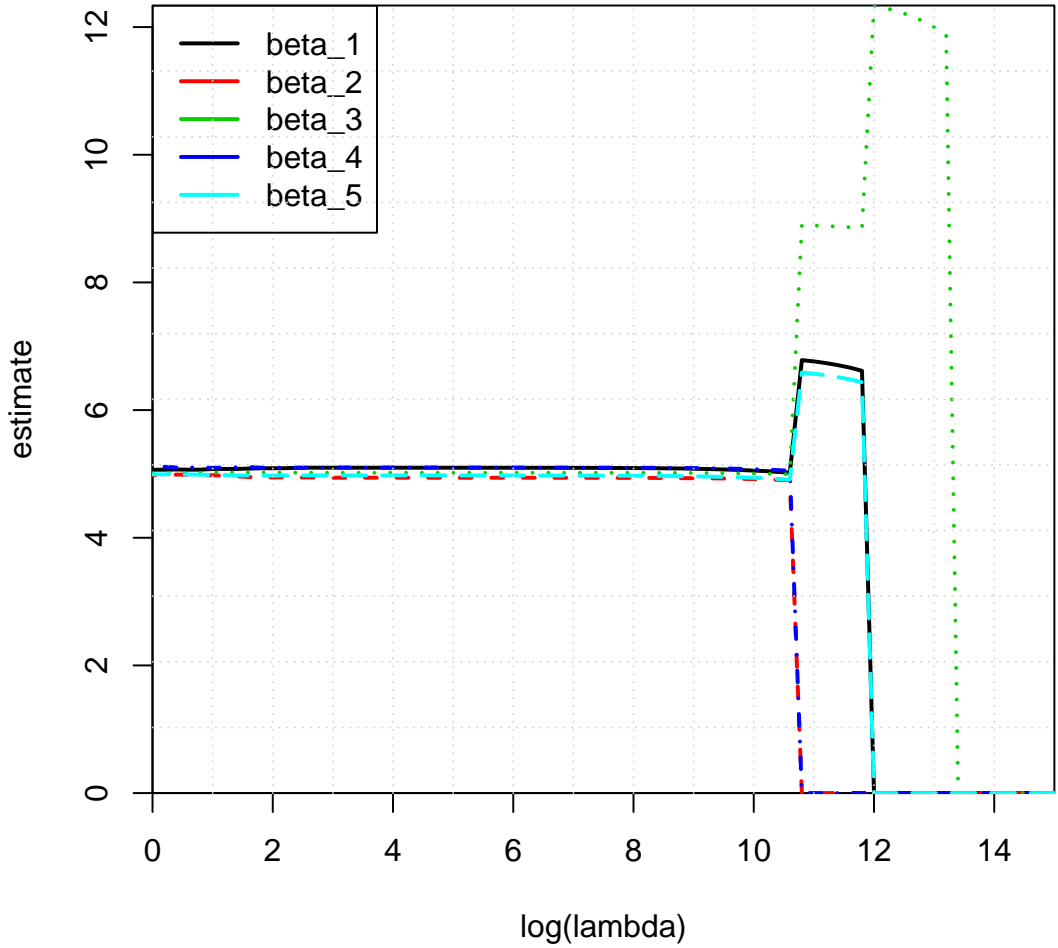}
\end{center}
\label{betahat}
\smallskip
\footnotesize 
Note: In this figure, we plot the regularization paths of ($\hat\beta_1$, $\hat\beta_2$,..., $\hat\beta_5$), where $N = 600, p=500$ and $\alpha = 0.1$.  Discontinuous points in the regularization path of a specific $\hat\beta_j$ can be explained by other $\hat\beta_i$'s dropping to 0, a special feature of $\ell_\alpha$ regularization caused by the discontinuity of the $\tau$ function at $h_{\lambda,\alpha}$. Because $x_1, x_2, ..., x_5$ are positively correlated, the drop of one $\hat\beta$ increases the estimated value of others. 
\end{figure}

The larger $\alpha$ is, the quicker $\hat\beta_1$ shrinks. This finding suggests the $\ell_\alpha$ estimator is less biased than the lasso estimator ($\alpha = 1$) regarding the true non-zeros, and we can further reduce estimation bias by choosing a smaller $\alpha$. Figure \ref{betahat} shows that when  $\hat\beta$ drops to 0 as $\lambda$ increases, the regularization paths of other $\hat\beta$'s are affected and immediately make jumps.

For a suitable range of $\lambda$, $\ell_\alpha$ regularization introduces more sparsity than lasso. The right column in Figure \ref{proxiS1} shows how the number of non-zero $\hat\beta_i$'s change with $\lambda$. We point out that, the path of $\ell_{0.1}$ drops most quickly when $\log(\lambda)$ is less than 0. Though the regularized models with this range of $\lambda$ are still redundant since number of non-zero $\hat\beta_i$'s is greater than the true value, $\ell_{\alpha}$ gives a more sparse model than lasso. 

Once the true model is achieved, $\ell_\alpha$ tends to stay on it though the regularization parameter $\lambda$ keeps increasing, especially when $\alpha$ is small. In other words, when $\lambda$ is small and there are many redundant variables in the model, the regularization of $\ell_\alpha$ is stronger than lasso; when $\lambda$ is large and we are close to the true model,  the regularization of $\ell_\alpha$ is weaker than lasso. In Figure \ref{proxiS1}, we see that when $\log(\lambda)$ is close to 0, the performances of $\ell_{0.1}$ and lasso are similar. This is where the relative strength of regularization gets reversed. The plots summarize the estimated regularization paths and indicate again that $\ell_\alpha$ regularization is less biased and more robust for variable selection. $\ell_\alpha$ regularization possesses better performances with a wider range of $\lambda$.

We also show the performance of $\ell_\alpha$ regularization under two different correlation strengths: Cor$(X_i, X_j) = \rho^{|i-j|/3}$, where $\rho=0.1, 0.6$, representing low and high correlation, respectively. In this example, the regularization parameter $\lambda$ is chosen using five-fold cross-validation. The sizes of the datasets are $N^{train} = 600, N^{test}=600$, and the total number of runs is 100. In Table \ref{bias-compare}, we list the average prediction RMSE (out of sample), and the bias/variance of coefficient estimates (in sample) of $\ell_\alpha$ with other linear methods, including lasso, OLS, and the elastic net. For the elastic net, we always use the penalty $\frac{1}{2}\Vert\beta\Vert_1 + \frac{1}{4}\Vert\beta\Vert_2^2$. The dataset dimensions are $p=50,100$,and $500$ as before. Lasso is a benchmark, and normalize results by their counterparts under lasso. The RMSE is simply calculated as
\begin{equation*}
\text{RMSE}(Y, \hat Y)=\sqrt{ \frac{1}{n}\sum_{i=1}^n (y_i -\hat y_i)^2}
\end{equation*}
where \((y_1, y_2, \ldots, y_n)\) are observed values and \( (\hat y_1,\hat y_2, \ldots,\hat y_n)\) are the predicted values.

We find that $\ell_\alpha$ gives more accurate predictions in all dimensional and correlation settings. By comparing the average RMSE, we see that small $\alpha$ produces small prediction errors. $\ell_\alpha$ also produces less bias than lasso and elastic net in almost all cases, though the estimator variances get larger for those non-zero estimates. For $\beta_6=0$, we also emphasize that in the high-correlation setting, $\ell_\alpha$ estimates $\beta_6$ with great accuracy.  Both $\ell_{0.1}$ and $\ell_{0.5}$ give the correct $\hat\beta_6 = 0$ with zero estimator variance. $\ell_\alpha$ performs extremely well when dealing with those redundant variables. Imagining $\ell_\alpha$ encourages sparser models than the lasso and elastic net is therefore not hard. This advantage of $\ell_\alpha$ regularization tends to stay even under a high-correlation and high-dimensional design setting.

\subsection{Predictive Performance }\label{empirical-application}
The purpose of this section is to compare the prediction performance of our $\ell_\alpha$ regularization and other commonly used machine-learning methods. We follow the literature of market-demand estimation, particularly \cite*{bajari2015machine, bajari2015demand}. Marketing researchers use measures such price elasticities and lift estimates to assess the effectiveness of promotions. $\ell_\alpha$ provides less biased estimates with a prediction error that is only slightly higher for those black-box nonlinear methods.  We compare the performance of the alpha norm with other widely used machine-learning methods: OLS, generalized linear model boosting (GLMBoosting), random forests, support vector machines (SVMs), lasso, ridge, and the elastic net.

GLMBoosting improves the predictive ability by iteratively reweighting the estimated regression and classification functions. Reducing the bias is profitable but may sacrifice some prediction accuracy as the trade-off. Random forests are the most popular machine-learning method. During the training process, the model constructs multiple decision trees and outputs the class that is the mode of the classes (classification) or mean prediction (regression) of the individual trees. This method is often time-consuming and overfits the data without proper trimming. A Support Vector Machine(SVM) map points into different categories in space, and choose the hyperplane that maximizes the distance between the nearest data points on each side. 

Adding machine-learning methods  checks the robustness of our results. Random forest can achieve better prediction in some cases, whereas $\ell_\alpha$ is easier to interpret. From a computational perspective, the $\ell_\alpha$ estimator takes a far shorter time in tuning and run-time relative to non-parametric tree methods.

\subsubsection{Discrete Choice Model}\label{set-up}
The goal is to analyze customers' preferences among products, where our products
are collectively exhaustive, mutually exclusive, and finite. Suppose we have $J$ products and observe their sales in $M$ markets, and each market has $N_m$ customers. Assuming each customer only chooses one product based on higher expected utility, the probability of
customer $n$ in market $m$ choosing product $j$ is defined as
\[P_{mnj}=\text{P}(\text{customer $n$ in market $m$ choose product $j$}).\]

\noindent Since each customer only chooses one product in each market, \(\sum_j P_{mnj}=1, 0 \leq P_{mnj} \leq 1, j=1,\ldots,J\).  The choice of product $j$ by person $n$ in
market m, denoted by  \(y_{mnj}\), is given by
\begin{equation*}
y_{mnj}=\left\{
\begin{array}{rl}
1, &  U_{mnj}>U_{mnk}, k\neq j\\
0, & otherwise.
\end{array}
\right.
\end{equation*}

\noindent Here \(U_{mnj}\) is the utility of product $j$ for customer $n$ in
market $m$:
\begin{equation}
U_{mnj}=\beta_{0}+X_{mj}\beta+\epsilon_{mnj},
\end{equation}

\noindent  where \(X_{mj}\) is a vector of characteristics of product $j$ in market $m$. The parameter  \(\beta\) is a set of parameters giving the effects of variables on probabilities in the market, and \(\epsilon_{mnj}\) captures the individual difference for product  $j$ in market $m$, specifically for person $n$. The choice probability is given by
\begin{align*}
P_{mnj}= &\text{P}(y_{mnj}=1)=\text{P}(\cap_{k\neq j} \{ U_{mnk}<U_{mnj}\})\\
= & \text{P}(\cap_{k\neq j} \{ \epsilon_{mnk}-\epsilon_{mnj}<X_{mj}\beta_{m}-X_{mk}\beta_{m}\}).
\end{align*}

 In our simulation, we primarily explore the case of one product. Then we have the choice of choosing the product and not choosing the product (option 1 vs option 0):
\begin{equation}
\left\{
\begin{array}{l}
U_{mn1} = \beta_0 + X_{m}\beta+\epsilon_{mn}\\
U_{mn0}=0\\
\epsilon_{mn} \sim \mbox{Logis}(0,1).
\end{array}
\right.
\end{equation}

\noindent  Here $U_{mn1}$ is the utility of this product for customer $n$ in market $m$. \(\beta\) describes how characteristics influence the
expected utility, and is also a vector
of length K, and it independently comes from a multivariate normal distribution. $\beta_0+X_{m}\beta$ is the systematic utility of this product in market $m$, and $\epsilon_{mn}$ is the random utility for customer $n$ in market $m$. Here $U_{mn0}$ is the utility of not choosing the product, and we set it to be 0. Here \(X_{m}\) is a vector of  $K$ product characteristics.

Then \(P_{mn1}\) have a type 1 extreme distribution as
\begin{equation}
P_{mn1}=\frac{\exp(\beta_0+X_{m}\beta)}{1+\exp(\beta_0+X_{m}\beta)}.
\end{equation}

For large enough $N$, the sample probability of
product 1 being chosen in market $m$ converges to this extreme distribution value. \cite*{bajari2015demand} construct a data generating process for each of the characteristics  \(X_{m}\). We simulate them independently and identically from multivariate log-normal distribution:
\begin{equation*}
X_{m}  \sim \mbox{logNormal}(0,\Sigma_{m}), \mbox{diag}(\Sigma_{m})  \sim \mbox{Unif}(0.5,1.5;K). 
\end{equation*}

\noindent  Next we add confounding variables. Given $K_c$ confounding variables here, and that they are weakly correlated with true variables, we are actually constructing
\begin{equation*}
(X_{m}, X^c_{m}) \sim \mbox{logNormal}(0,\Sigma ), \Sigma=\left[
\begin{array}{cc}
\Sigma_{m} & C_1\\
C_1^T & \Sigma_{m}^c
\end{array}\right],
\end{equation*}

\noindent  where $X^c_{m}$ is the matrix of confounding variables, $C_1$ is the covariance matrix of $X_{m}$ and $X^c_{m}$, and we control their correlation to be weak. $\Sigma^c_{m}$ is the covariance matrix of $X_{m}$.  To control the correlation among true variables and confounding variables, we construct the big covariance matrix $\Sigma$ as
\begin{equation}\label{Sigma_Construct}
[\Sigma]_{i,j}=\rho^{\frac{|i-j|}{3}}.
\end{equation}

\noindent  Here $\rho$ is a pre-set correlation parameter, $0<\rho<1$. Combining $X_{m}$ and $X^c_{m}$ gives us the total characteristics of the product in market $m$: $X^t_{m}=(X_{m}, X^c_{m}) $:

Then we add categorical variables
into the datasets. Specifically, we create
binary variables by setting a cutoff value $T$ and let
\begin{equation*}
\tilde{X}_{mk}=\left\{
\begin{array}{ll}
1,& X_{mk}>T\\
0, & otherwise.
\end{array}
\right.
\end{equation*}

\noindent The sample share of the product in market $m$ is defined as
\begin{equation}
S_{m}=\frac{1}{N_m}\sum_{n=1}^{N_m} \mathds{1}_{\{U_{mn1}>0\}}.
\end{equation}

\noindent  Here $N_m$ is the number of customers in market $m$. To make the case simpler, we assume the number of customers in each market is the same. If we set $N_m$ large enough,  $S_m$ follows the extreme value distribution and converges to the real share. Following our empirical analysis in Section \ref{empirical-analysis}, instead of using sample shares as a response, we model the log of the quantity(unit) sold in each market:
\begin{equation*}
\mbox{log}(Q_m)=\mbox{log}(\sum_{n=1}^{N_m} \mathds{1}_{\{U_{mn1}>0 \}}),
\end{equation*}

 \noindent  where \(\mbox{log}(Q_m)=(\mbox{log}(Q_1), \mbox{log}(Q_2),\ldots,\mbox{log}(Q_M))'\) is the vector of the log of units of this product sold in each market.
\(X_{m}=(X^t_{1},X^t_{2},\ldots,X^t_{M})'\) is this product's characteristics in each market.

\subsubsection{Empirical results: Model comparison}\label{comparision-of-model}

To estimate and perform model comparison, we partition the data into two
parts: training \(\mbox{DGP}^{(1)}\) and testing \(\mbox{DGP}^{(2)}\). And we control \(\mbox{DGP}^{(1)}\)  and \(\mbox{DGP}^{(2)}\) to be the same size. For \(\mbox{DGP}^{(1)}\), we use this part of the data to estimate the best tuning
parameters for SVM, GLMBoosting, and random forests. We  find the best $\hat \lambda$ and best $\hat \alpha$ for the alpha norm via a five-fold cross-validation, where $\alpha $ is chosen from $\{0.1,0.5,0.9\}$. We also use five-fold cross-validation to choose the best $\hat \lambda$ for lasso, ridge, and the elastic net. We plug in the tuning parameters to construct all models \(\{f_i\}\).

For \(\mbox{DGP}^{(2)}\), we obtain \(\hat{y_i}=\text{predict}(f_i,\mbox{DGP}^{(2)})\) for each method, and we estimate their corresponding out-of-sample RMSE. 
%In addition, we add an oracle model in our comparison table, in which we use OLS to regress $\mbox{log}(Q_m)$ on the true predictors, because we already know them from constructing the datasets. Therefore, this oracle model should be the true model and have the smallest out-of-sample RMSE among all linear models. We can compare the performance of the $\ell_\alpha$ and oracle model to see how close we come to fitting.

As our panel data in Section \ref{empirical-analysis} only have two continuous predictors, we include two continuous predictors in the true-predictor set.  We try two different true-predictor cases. In the first instance, we include two continuous predictors and two binary predictors as true predictors. We estimate the RMSE of all models in different situations and calculate their RMSE ratios with the alpha norm as the benchmark. Table \ref{ratio_2_2} displays the results. Similarly, in the second case, we use two continuous and 20 binary predictors as true predictors, and present the results in Table \ref{ratio_2_20}. To further explain our data-generating process, for the rest of the redundant predictors, we control their correlation with the true predictors via $\rho$ in equation \ref{Sigma_Construct}. In the low-correlation case, we let $\rho=0.1$,whereas in the high-correlation case, we make $\rho=0.6$. The confounding predictors are 50\% numerical and 50\% categorical. We want to show our $\ell_\alpha$ method outperforms other linear methods in a high-correlation and high-dimensional case, by altering correlation $\rho$,  market size $M$, and the number of predictors in total $K$.

From the results in  Table \ref{ratio_2_2} and Table \ref{ratio_2_20}, we  see that most of the entries are larger than 1, and ratios tend to be greater in the high-correlation case, or in the high-dimensional case, where $K$ is equal to or greater than $M$. In most cases, our $\ell_\alpha$ method outperforms the other methods except the random forest. The alpha norm can provide proper model selection and marginal effect estimates, however, whereas the random forest is difficult to interpret and lacks flexibility when we want to define subgroups. 

%%%%%%%%%%%%%%%%%%%%%%%%%%%%%%%%%%%%%%%%%
\section{Store-Level Market Demand Data}\label{empirical-analysis}

\subsection{Sales Prediction}\label{sales_prediction}

The marketing example is about grocery store sales for salted snacks. Our dataset uses scanner-panel data on grocery stores from IRI Marketing Research. A unit of observation is product $j$, uniquely defined by a UPC (Universal Product Code), in store $m$ in week $t$. The number of observations is
15,339, which includes 100 unique products. 

Let \(q_{jmt}\) be the
number of bags of salty snack j sold in store $m$ in week $t$. If
\(q_{jmt} = 0\), the possible situation can be zero sale, being out of stock
or missing observation. The price \(p_{jmt}\) is defined
as the quantity-weighted average of prices for product $j$ in store $m$ in
week $t$. Therefore, if \(q_{jmt}\) = 0, the weight is also set to zero. The general regression model is of the form
\begin{equation*}
Y_{jmt}=f(X_{\mbox{price},jmt}, X_{\mbox{product},jmt},X_{\mbox{promotion},jmt},X_{\mbox{week},t})+\epsilon_{jmt}.
\end{equation*}

\noindent Table \ref{meaning-variables} provides detailed summary information on our predictors. In our model, we do not use the variable iri\_key, which is a unique label for each store, because it is severely unbalanced in the data.  The weeks are transformed into a combination of year and week. Specifically, we use six years and 52 weeks to present the 300 weeks in our dataset to deal with the unbalanced data issue.

For model validation, again we randomly separate the dataset into two parts: training $\mbox{DGP}^{(1)}$ and testing $\mbox{DGP}^{(2)}$. Then we use $\mbox{DGP}^{(1)}$ to estimate the best tuning parameters and construct the model and apply the models to $\mbox{DGP}^{(2)}$ to evaluate the out-of-sample RMSE.

In total our dataset contains 15 predictors with three continuous variables: quantity (number of units sold), price, and equivalized volume (or all commodity volume), which represent the store revenue from all products sold. Quantity ranges from 1 to 1,505 with mean 16.36 and standard deviation 40.37. Price ranges from 0.2 to 9.99 with mean 2.11 and standard deviation 1.003. Equivalized volume ranges from 0.0469 to 3 with mean 0.5139 and standard deviation 0.266.

Table \ref{market_rmse} provides out-of-sample RMSE ratios for the $\ell_\alpha$ estimator as the benchmark and $R^2$ of all models . Out-of-sample $R_{\mbox{OOS}}^2$ is simply calculated as
\begin{equation}\label{eq:OOS_Rsquare}
R^2_{\mbox{OOS}} = 1- \frac{\sum (\hat{y_i}-y_i)^2}{\sum (y_i -\bar y)^2},
\end{equation}

\noindent where $\hat {y_i}$ and $y_i$ are the predicted and observed values, respectively. The $\ell_\alpha$ regularization beats all other linear methods, with a smaller RMSE and larger $R^2$. The final solution selects 158 predictors.

\begin{figure}[!ht]
\caption{Out-of-Sample RMSE and Number of Predictors under Different Models}\label{fig:15000}
\begin{center}
\subfigure[RMSE]{
\includegraphics[width=0.5\paperwidth]{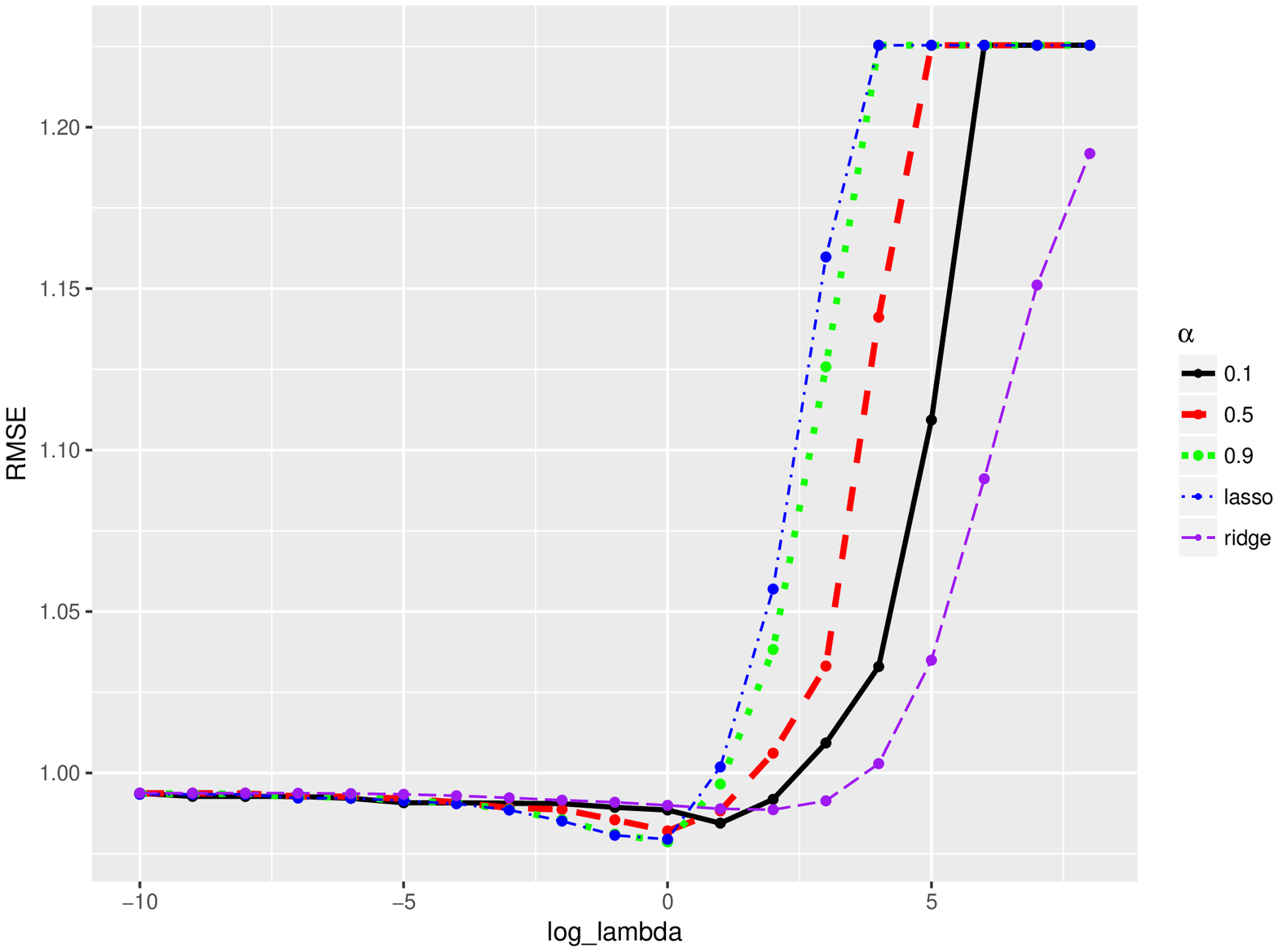}
}\label{fig:MSE_15000}
\subfigure[Number of predictors selected]{
\includegraphics[width=0.5\paperwidth]{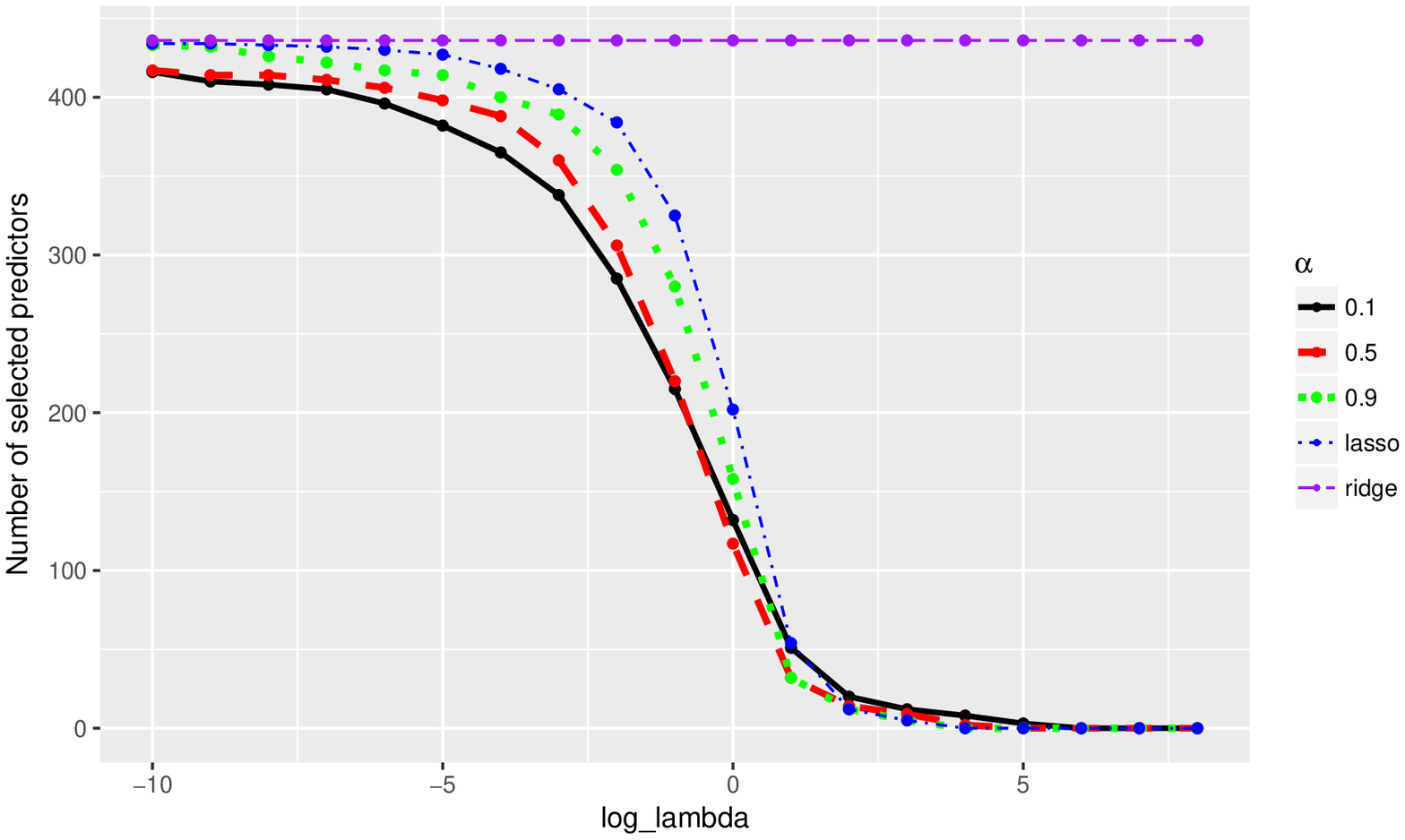}
}\label{fig:N_select_15000}
\end{center}
\smallskip
\footnotesize
 Note: Figure (a) shows how out-of-sample RMSE changes when lambda increases. Figure (b) shows the number of selected predictors selected. We can see that when the $\lambda$ is growing, the penalty terms in these methods are also growing, which encourages a sparser model with fewer predictors. The alpha norm removes predictors faster than lasso when $\lambda$ is small and removed predictors are often useless,  but slows down the speed when $\lambda$ is larger and removed predictors are more significant.

\end{figure}

To better understand the process of regularization, we plot the trend of changes in RMSE, and the number of predictors of the $\ell_\alpha$ and ridge regression when we change lambda in Figure \ref{fig:15000}. (Notice we do not draw this plot with lasso from {\tt glmnet}; instead, we use the alpha norm with \(\alpha=1\).)

The top predictors of those regularized methods can be found from the penalization path in Figure \ref{fig:15000}. The top predictors we can extract from the path include price, equivalized volume, promotion, brands (e.g., Lays and Ruffles), flavors (e.g., original, classical, regular, and sour cream and onion), the cooking method (e.g.,kettle-cooked), and fat content (e.g., reduced fat). Among these top predictors, many strands of literature have discussed the effect of price, promotion, brand, and equivalized volume. % More research to be implemented  
Our method, however, provides an even closer look at the incremental effect of a particular brand, flavor, cooking method, and fat content, which can help grocery stores develop detailed strategies to improve their inventory. For example, a larger proportion of Lays snacks stock leads to higher sales.

\subsection{Promotion Lift Estimates}\label{lift estimate}

Section \ref{sales_prediction} shows promotion is always selected as a top predictor. Now we estimate the lift of a promotion. The products in our dataset are tagged with a promotion label if their price deduction is greater than 5\%. We want to see the incremental effect on sales generated by promotion (see \cite*{bajari2015demand}). %\cite*{chintagunta2011structural} provide a nonparametric estimation method that combines with a source of exogenous variation to address issues related to functional form and nonrandom variations.

First, we split the data into a training set $\mbox{DGP}^{(1)}$ and testing set $\mbox{DGP}^{(2)}$. $\mbox{DGP}^{(1)}$ contains all the records with no promotion, and $\mbox{DGP}^{(2)}$ contains all the records with promotion. We find that we have 11,348 non-promotion records, which account for 74\% of our data. We use $\mbox{DGP}^{(1)}$ to train all models and provide predictions for $\mbox{DGP}^{(2)}$. Then we calculate the lift factor:
\begin{equation*}
\mbox{Lift factor}=\frac{\mbox{Actual sales}}{\mbox{Baseline sales}}=\frac{y}{\hat y}.
\end{equation*}

We use the predictions from the non-promotion model as the baseline sales and compare them to the actual sales. The incremental effect on the lift factor is defined by
\begin{equation*}
\Delta Q =\mbox{Lift factor}\cdot \mbox{Baseline sales}- \mbox{Baseline sales}=(\mbox{Lift factor}-1)\cdot \mbox{Baseline sales}.
\end{equation*}

When we predict the sales using a log-linear model of the form:
\begin{align*}
&\log(Q)=\alpha-\eta \log(P)+\sum_{i=1}^p \beta_i X_i,\\
&\log(Q')=\alpha-\eta \log((1-\gamma)P)+\sum_{i=1}^p \beta_i X_i +\beta_{\mbox{prom}} \mbox{Prom}.
\end{align*}

\noindent Here, $Q$ is the quantity of products in the absence of promotion, and $X_i$ is all the predictors. $P$ is the price of the product, $Q'$  is the quantity of products when promotion is considered, and Prom is the dummy indicator for promotion. $\beta_{\mbox{prom}} $ is the corresponding coefficient and $\gamma$ is the discount applied to price in the promotion, and in our model, $\gamma>0.05$.

The lift factor is calculated from the comparison of the two models as
\begin{align*}
\mbox{log(Lift)}&=log(Q')-log(Q)=-\eta log(1-\gamma)+\beta_{\mbox{prom}}
\mbox{Prom},\\
\mbox{Lift factor}&=\frac{Q'}{Q}=\mbox{exp}(-\eta log(1-\gamma)+\beta_{\mbox{prom}} . {\mbox{Prom}} ).
\end{align*}

\begin{figure}[!ht]
\caption{Distribution of Log(Lift)}\label{fig:dist_log_Lift}
\begin{center}
\includegraphics[width=0.5\paperwidth]{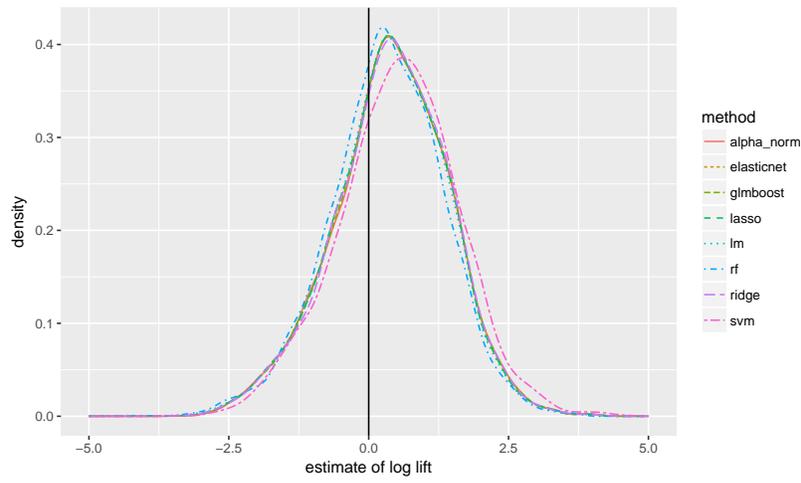}
\end{center}
\smallskip
\footnotesize
Note: We plot the distribution of estimated log lift of each model here. We can see the shapes of different models are quite similar. Though the means are obviously positive, the distributions all have a small proportion of negative values, which indicates that in some cases, the promotion may not have an incremental effect on sales.
\end{figure}

\noindent  The average of realized lift factors can predict the future lift from a promotional event. Figure \ref{fig:dist_log_Lift} plots the distribution of log(lift) calculated under different models. The distributions generated from different models are similar, and the mean of each distribution is positive as expected.

Figure \ref{fig:dist_log_Lift} shows that a large number of models have negative estimates for log(Lift), though the average effect of promotion is positive and statistically significant. We want to investigate what may play a significant role in deciding the magnitude and sign of the promotional effect. Finally, models of log(lift) are helpful in determining promotional strategies concerning an individual product.

\begin{figure}[!ht]
\caption{Accuracy of Log(Lift) estimate}\label{fig:ols_pr_boot}
\begin{center}
\includegraphics[width=0.5\paperwidth]{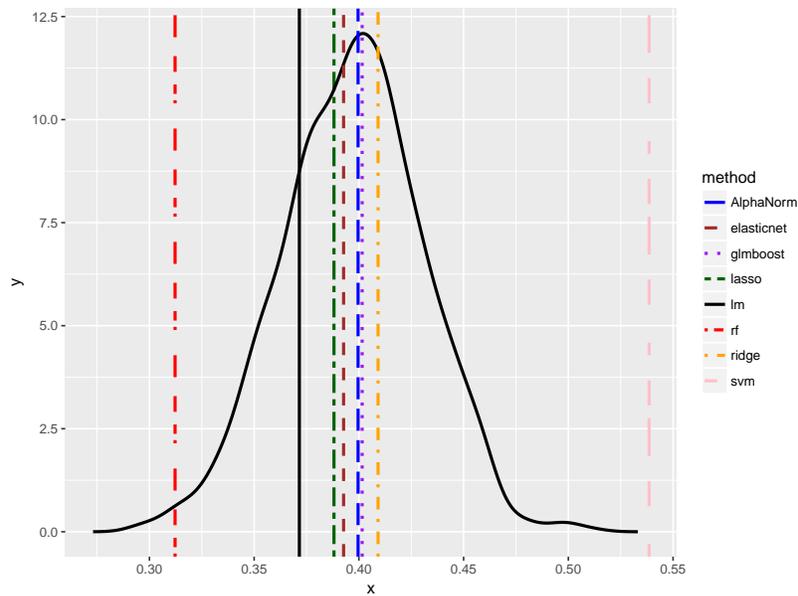}
\end{center}
\smallskip
\footnotesize
Note: We generate this plot using 1,000 resampling bootstraps of the OLS model as in Section \ref{sales_prediction}. We can see the distribution of $\beta_{\mbox{promotion}}$ in the OLS model is close to Gaussian with an average of  0.4. The mean of estimated lift given by $\ell_\alpha$ model is close to 0.4, which is expected to be much less biased than svm and rf.. The vertical lines are the mean of estimated log lift in Section \ref{lift estimate}, and except for the SVM and random forests, they appear to be close to the expectation given by the OLS model.
\end{figure}

Our lift estimates based on the $\ell_\alpha$ estimation are less biased than linear regression methods. We use resampling bootstrap to get the distribution of $\beta_\text{Prom}$ from OLS estimate, and we hope it will approximate to the true distribution since OLS gives unbiased estimates. For each resampling model, we randomly select half of the observations and use them to fit the OLS. We plot the distribution of $\beta_\text{Prom}$ in Figure \ref{fig:ols_pr_boot}. The plot of log(lift) estimated by the alpha norm is very close to the mean of OLS bootstrapping. And it is expected to be less biased than rf and svm. All methods give positive average lifts, with positive and negative increments in the estimate. The large variance in the estimation suggests the negative lifts come from variance rather than bias, see \cite{bajari2015demand}. Further improvements in lift estimates occur if we use the variables selected in lasso or the alpha norm.

%Finally, we regress log(lift) on all non-promotion predictors, and we can track the top predictors from the regularization path. Notice we use the estimated log(lift) as our response in the model, which is very different from the previous model. The model selects price, brand, and flavor as important in estimating the incremental effect of promotion.  RMSE ratios are compared to the alpha norm, and out-of-sample $R^2$ (using equation \ref{eq:OOS_Rsquare}) of all models are shown in Table \ref{log_Lift_est}. 

%%%%%%%%%%%%%%%%%%%%%%%%%%%%%%%%%%%%%%%%%
\section{Discussion}\label{discussion}
$\ell_\alpha$ regularization provides a useful tool for predictor selection in marketing and economics. Scanner-panel data usually have thousands of binary dummy variables, and many of the predictors do not predict sales; hence, variable selection is needed. Our $\ell_\alpha$ regularization can jump to a sparse solution.  Post-lasso variables can increase the fit of the model in high-dimensional sparse cases. For the applications we use here, $\ell_\alpha$ regularization finds a better solution to in-sample overfitting and is more adaptive than lasso, as the degree of the norm can be chosen from 0 to 1, thus applying to both sparse and extreme sparse models. 

Our empirical analysis shows $\ell_\alpha$ regularization does improve predictions versus traditional linear regression and machine-learning black-box techniques.  Predicted sales and selected variables can be used for inventory planning and can predict the outcome of potential marketing strategies. Also, the alpha norm can be particularly useful when the practitioners want to study the significant predictors for a particular subpopulation. Our alpha norm can efficiently shrink the predictor size and pick significant predictors according to the features of the response.  

In contrast to machine-learning approaches, the $\ell_\alpha$ regularization benefits from interpretability of the marginal effects. The impact of a particular product or flavor can be assessed. $\ell_\alpha$ regularization is more efficient in estimating the model than nonlinear methods while providing relatively similar performance. A further extension could be to apply our methodology to demand and supply-demand estimation as in \cite*{brian2015bayesian}, when targeting a specific incremental effect, such as the lift of promotion. Similar approaches can be applied to estimate lift of in-store displays, which usually have multiple levels. 

\clearpage
\bibliography{alphanorm}

\begin{thebibliography}{}

\bibitem[\protect\astroncite{Allenby et~al.}{2004}]{allenby2004choice}
Allenby, G.~M., Shively, T.~S., Yang, S., and Garratt, M.~J. (2004).
\newblock A choice model for packaged goods: Dealing with discrete quantities
  and quantity discounts.
\newblock {\em Marketing Science}, 23(1):95--108.

\bibitem[\protect\astroncite{Andrews et~al.}{2011}]{andrews2011comparison}
Andrews, R.~L., Currim, I.~S., and Leeflang, P.~S. (2011).
\newblock A comparison of sales response predictions from demand models applied
  to store-level versus panel data.
\newblock {\em Journal of Business \& Economic Statistics}, 29(2):319--326.

\bibitem[\protect\astroncite{Angrist}{2001}]{angrist2001estimation}
Angrist, J.~D. (2001).
\newblock Estimation of limited dependent variable models with dummy endogenous
  regressors: simple strategies for empirical practice.
\newblock {\em Journal of Business \& Economic Statistics}, 19(1):2--28.

\bibitem[\protect\astroncite{Attouch et~al.}{2013}]{attouch2013convergence}
Attouch, H., Bolte, J., and Svaiter, B.~F. (2013).
\newblock Convergence of descent methods for semi-algebraic and tame problems:
  proximal algorithms, forward--backward splitting, and regularized
  gauss--seidel methods.
\newblock {\em Mathematical Programming}, 137(1-2):91--129.

\bibitem[\protect\astroncite{Bajari et~al.}{2015a}]{bajari2015machine}
Bajari, P., Nekipelov, D., Ryan, S., and Yang, M. (2015a).
\newblock Machine learning methods for demand estimation.
\newblock {\em American Economic Review: Papers \& Proceedings},
  105(5):481--485.

\bibitem[\protect\astroncite{Bajari et~al.}{2015b}]{bajari2015demand}
Bajari, P., Nekipelov, D., Ryan, S.~P., and Yang, M. (2015b).
\newblock Demand estimation with machine learning and model combination.
\newblock Technical report, National Bureau of Economic Research.

\bibitem[\protect\astroncite{Bredies et~al.}{2015}]{bredies2015minimization}
Bredies, K., Lorenz, D.~A., and Reiterer, S. (2015).
\newblock Minimization of non-smooth, non-convex functionals by iterative
  thresholding.
\newblock {\em Journal of Optimization Theory and Applications},
  165(1):78--112.

\bibitem[\protect\astroncite{Briesch et~al.}{2010}]{briesch2010nonparametric}
Briesch, R.~A., Chintagunta, P.~K., and Matzkin, R.~L. (2010).
\newblock Nonparametric discrete choice models with unobserved heterogeneity.
\newblock {\em Journal of Business \& Economic Statistics}, 28(2):291--307.

\bibitem[\protect\astroncite{Fong et~al.}{2010}]{fong2010private}
Fong, N.~M., Simester, D.~I., and Anderson, E.~T. (2010).
\newblock Private label vs. national brand price sensitivity: Evaluating
  non-experimental identification strategies.
\newblock Technical report, Massachusetts Institute of Technology.

\bibitem[\protect\astroncite{Friedman
  et~al.}{2010}]{friedman2010regularization}
Friedman, J., Hastie, T., and Tibshirani, R. (2010).
\newblock Regularization paths for generalized linear models via coordinate
  descent.
\newblock {\em Journal of Statistical Software}, 33(1):1--22.

\bibitem[\protect\astroncite{Hartmann et~al.}{2011}]{hartmann2011identifying}
Hartmann, W., Nair, H.~S., and Narayanan, S. (2011).
\newblock Identifying causal marketing mix effects using a regression
  discontinuity design.
\newblock {\em Marketing Science}, 30(6):1079--1097.

\bibitem[\protect\astroncite{Marjanovic and
  Solo}{2012}]{marjanovic2012optimization}
Marjanovic, G. and Solo, V. (2012).
\newblock On $l_q$ optimization and matrix completion.
\newblock {\em IEEE Transactions on signal processing}, 60(11):5714--5724.

\bibitem[\protect\astroncite{Marjanovic and Solo}{2013}]{marjanovic2013exact}
Marjanovic, G. and Solo, V. (2013).
\newblock On exact $l_q$ denoising.
\newblock In {\em In Acoustics, Speech and Signal Processing (ICASSP),2013 IEEE
  International Conference}, pages 6068--6072. IEEE, New York.

\bibitem[\protect\astroncite{Marjanovic and
  Solo}{2014}]{marjanovic2014sparsity}
Marjanovic, G. and Solo, V. (2014).
\newblock $l_q$ sparsity penalized linear regression with cyclic descent.
\newblock {\em IEEE Transactions on Signal Processing}, 62(6):1464--1475.

\bibitem[\protect\astroncite{Mazumder et~al.}{2011}]{mazumder2012sparsenet}
Mazumder, R., Friedman, J.~H., and Hastie, T. (2011).
\newblock Sparsenet: Coordinate descent with nonconvex penalties.
\newblock {\em Journal of the American Statistical Association},
  106(495):1125--1138.

\bibitem[\protect\astroncite{Mullainathan and
  Spiess}{2017}]{mullainathan2017machine}
Mullainathan, S. and Spiess, J. (2017).
\newblock Machine learning: an applied econometric approach.
\newblock {\em Journal of Economic Perspectives}, 31(2):87--106.

\bibitem[\protect\astroncite{Polson et~al.}{2015}]{polson2015proximal}
Polson, N.~G., Scott, J.~G., and Willard, B.~T. (2015).
\newblock Proximal algorithms in statistics and machine learning.
\newblock {\em Statistical Science}, 30(4):559--581.

\bibitem[\protect\astroncite{Polson and Sun}{2017}]{polson2017bayesian}
Polson, N.~G. and Sun, L. (2017).
\newblock Bayesian $ l_0 $ regularized least squares.
\newblock Technical report, University of Chicago.

\bibitem[\protect\astroncite{Varian}{1982}]{varian1982nonparametric}
Varian, H.~R. (1982).
\newblock The nonparametric approach to demand analysis.
\newblock {\em Econometrica: Journal of the Econometric Society},
  50(4):945--973.

\bibitem[\protect\astroncite{Viard et~al.}{2015}]{brian2015bayesian}
Viard, V.~B., Gron, A., and Polson, N.~G. (2015).
\newblock Bayesian estimation of nonlinear equilibrium models with random
  coefficients.
\newblock {\em Applied Stochastic Models in Business and Industry},
  31(4):435--456.

\bibitem[\protect\astroncite{Zeng et~al.}{2014}]{zeng2014cyclic}
Zeng, J., Peng, Z., Lin, S., and Xu, Z. (2014).
\newblock A cyclic coordinate descent algorithm for $l_q$ regularization.
\newblock Technical report, Xi'an Jiaotong University.

\end{thebibliography}

%%%%%%%%%%%%%%%%
% Appendix
%%%%%%%%%%%%%%%%%%%%%%%%%%%%%%%%%%%%%%%%%
\clearpage
\section*{Appendix}
\appendix
\section{$\ell_\alpha$ Regularization Algorithm}\label{appendix}
 
Specifically, in \cite*{marjanovic2014sparsity}, the algorithm starts with an initial estimate of $\beta$, for example, $\hat\beta^{ols}$ or 0, which is denoted as $\beta^1$.  Then the initial residual is calculated as $r^1:=y-X\beta^1$. Let $k$ be the iteration counter and let $i$ be the coefficient coordinate to be updated in the $k$-th iterate. Because $\beta\in\mathds{R}^p$, we simply have $i = p$ when  $0 \equiv k \text{ mod } p$, and $i = k \text{ mod } p$ otherwise. A complete iterate includes the calculation of adjusted gradient $z_i^k$, a proximal map, and the update of $\beta_i^k$. 

\begin{enumerate}[a)]
\item Calculate the adjusted gradient $z_i^k:=z(\beta_{-i}^k)$ for $i \in \{1,2,\ldots, p\}$ by
\begin{equation}
z_i^k=x_i^Tr^k+\beta_i^k.
\end{equation}

This expression of $z_i^k$ follows easily from the definition of $z^k$ and $r^k$. Here, $||x_i||_2 = 1$ , $\forall i $.

\item Using $z_i^k$ and the $\ell_\alpha$ optimization, calculate
the map

\begin{equation*}
\mathcal{T}(z_i^k,\beta_i^k) := 
\begin{cases}
\tau(z_i^k), &\text{if } |z_i^k|\neq h_{\lambda,\alpha} \\
\text{sgn}(z_i^k)b_{\lambda,\alpha}, &\text{if } |z_i^k|=h_{\lambda,\alpha},  \text{ and } \beta_i^k \neq 0\\
0, &\text{otherwise}, 
\end{cases}
\end{equation*}

when $|z_i^k|=h_{\lambda,q}$, $\tau(z_i^k) = \{0,\text{sgn}(z_i^k)b_{\lambda,\alpha}\}$ is implied.
\item Update $\beta^k$ by
\begin{equation}
\beta^{k+1} = \beta_{-i}^k+\mathcal{T}(z_i^k,\beta_i^k) e_i.
\end{equation}
Then $J(\beta^{k+1})\leq J(\beta^{k})$ by the definition of $\mathcal{T}(\cdot)$ and (\ref{descent}).

\item Update the residual:
\begin{eqnarray}
r^{k+1} &=& y-X\beta^{k+1} \nonumber\\
&=& y - X\left(\beta_{-i}^k+\beta_i^{k+1}e_i\right)\nonumber\\
&=& \left(y - X\beta_{-i}^k - \beta_i^k x_i\right) - \beta_i^{k+1}Xe_i + \beta_i^k x_i\nonumber\\
&=& r^k - \left(\beta_i^{k+1}-\beta_i^k\right)x_i.
\end{eqnarray}

\item Update the iteration counter $k$ by $k=k+1$.    
\item Update $i$ by $i = p$ when  $0 \equiv k \text{ mod } p$, and $i = k \text{ mod } p$ otherwise.
\end{enumerate}

%%%%%%%%%%%%%%%%%
% Table and Figure
\clearpage
\begin{table}[!ht]
\caption{Comparison of Different Linear Methods}
\label{bias-compare}
\vspace{-0.2in}
\begin{center}
\begin{tabular}{@{}cc|cccccc@{}}
\toprule
P                    & \multicolumn{1}{l|}{Measure}                            & $\ell_{0.1}$ & $\ell_{0.5}$ & $\ell_{0.9}$ & lasso & ols       & elastic net \\ \midrule
& \\
                     & \multicolumn{1}{l|}{}                                 & \multicolumn{6}{c}{low correlation}                                    \\
& \\
\multirow{5}{*}{50}  & \multicolumn{1}{l|}{RMSE}                             & 0.99      & 0.99      & 0.99      & 1 & 1.03     & 1.01       \\
                     & \multicolumn{1}{l|}{Bias $\beta_1$}   & 0.46      & 0.39      & 0.68      & 1 & -0.12    & 1.14       \\
                     & \multicolumn{1}{l|}{Var $\beta_1$} & 1.29      & 1.18      & 1.04      & 1 & 0.95     & 1.14       \\
                     & \multicolumn{1}{l|}{Bias $\beta_6$}   & 0.23      & 0.20      & 0.32      & 1 & -0.07    & 1.28       \\
                     & \multicolumn{1}{l|}{Var $\beta_6$} & 0.35      & 0.40      & 0.30      & 1 & 10.68    & 1.33       \\
                     &                                                       &            &            &            &       &           &             \\
\multirow{5}{*}{100} & \multicolumn{1}{l|}{RMSE}                             & 0.99     & 0.99     & 0.99     & 1 & 1.08     & 1.01      \\
                     & \multicolumn{1}{l|}{Bias $\beta_1$}   & 0.39     & 0.34     & 0.78      & 1& 0.01    & 1.07       \\
                     & \multicolumn{1}{l|}{Var $\beta_1$} & 1.05      & 1.08      & 1.02     & 1 & 0.78     & 0.93       \\
                     & \multicolumn{1}{l|}{Bias $\beta_6$}   & -0.23     & -0.19     & -0.01     & 1 & -1.80    & 1.63       \\
                     & \multicolumn{1}{l|}{Var $\beta_6$} & 0.64      & 0.41      & 0.08     & 1 & 35.09    & 2.82       \\
                     &                                                       &            &            &            &       &           &             \\
\multirow{5}{*}{500} & \multicolumn{1}{l|}{RMSE}                             & 0.98      & 0.98      & 0.99      & 1 & 2.42  & 1.01       \\
                     & \multicolumn{1}{l|}{Bias $\beta_1$}   & 0.36      & 0.28      & 0.67      & 1 & 0.12  & 1.04       \\
                     & \multicolumn{1}{l|}{Var $\beta_1$} & 1.32      & 1.06      & 0.95      & 1 & 5.67  & 1.08       \\
                     & \multicolumn{1}{l|}{Bias $\beta_6$}   & -0.69     & 0.00      & -0.11     & 1 & 11.44  & 4.91       \\
                     & \multicolumn{1}{l|}{Var $\beta_6$} & 3.49      & 0.00      & 0.10      & 1 & 6.28E+02  & 10.85      \\
                     &                                                       &            &            &            &       &           &             \\
                     & \multicolumn{1}{l|}{}                                 & \multicolumn{6}{c}{high correlation}                                   \\
                     &                                                       &            &            &            &       &           &             \\
\multirow{5}{*}{50}  & \multicolumn{1}{l|}{RMSE}                             & 0.99      & 0.99      & 1.00     & 1 & 1.03     & 1.00       \\
                     & \multicolumn{1}{l|}{Bias $\beta_1$}   & 1.13      & 0.64      & 0.71      & 1 & 0.04     & 0.92       \\
                     & \multicolumn{1}{l|}{Var $\beta_1$} & 0.98      & 1.36      & 1.09      & 1 & 0.91     & 1.02       \\
                     & \multicolumn{1}{l|}{Bias $\beta_6$}   & 0.00      & 0.00      & 0.08      & 1 & 0.65     & 3.76       \\
                     & \multicolumn{1}{l|}{Var $\beta_6$} & 0.00      & 0.00      & 0.04      & 1 & 17.66    & 4.18       \\
                     &                                                       &            &            &            &       &           &             \\
\multirow{5}{*}{100} & \multicolumn{1}{l|}{RMSE}                             & 0.99      & 0.99      & 1.00      & 1 & 1.08     & 1.00      \\
                     & \multicolumn{1}{l|}{Bias $\beta_1$}   & 0.97      & 0.81      & 0.79      & 1 & -0.08    & 0.84       \\
                     & \multicolumn{1}{l|}{Var $\beta_1$} & 0.85      & 1.37      & 1.07      & 1 & 0.76     & 0.94       \\
                     & \multicolumn{1}{l|}{Bias $\beta_6$}   & 0.00      & 0.00      & 0.06      & 1 & 1.79     & 3.87       \\
                     & \multicolumn{1}{l|}{Var$\beta_6$} & 0.00      & 0.00      & 0.02      & 1 & 17.67    & 5.05       \\
                     &                                                       &            &            &            &       &           &             \\
\multirow{5}{*}{500} & \multicolumn{1}{l|}{RMSE}                             & 0.99      & 0.99      & 0.99      & 1 & 2.41  & 1.01       \\
                     & \multicolumn{1}{l|}{Bias $\beta_1$}   & 0.74     & 0.25     & 0.61      & 1 & -6.56E-03 & 1.10       \\
                     & \multicolumn{1}{l|}{Var $\beta_1$} & 0.89      & 1.37      & 1.06      & 1 & 5.54  & 1.07       \\
                     & \multicolumn{1}{l|}{Bias $\beta_6$}   & 0.00      & 0.00      & 0.00      & 1 & -10.80 & 6.96      \\
                     & \multicolumn{1}{l|}{Var $\beta_6$} & 0.00      & 0.00      & 0.00      & 1 & 2.94E+02  & 12.52      \\ \bottomrule
\end{tabular}
\end{center}
\footnotesize
Note: $N^{\mbox{train}} = 600, N^{\mbox{test}}=600$ and the total number of runs is 100. Cor$(X_i, X_j) = \rho^{|i-j|/3}$ where $\rho=0.1, 0.6$. The regularization parameter $\lambda$ is chosen 5-fold cross-validation. We use lasso as a benchmark and divide all the result values by their counterparts given by lasso. RMSE is the out-of-sample performance in the test sample (averaged by 100 runs). Bias and variance are the in-sample performance in the training sample.
\end{table}

\begin{table}[!ht]
	\caption{Model Comparison for Discrete-Choice Models I}\label{ratio_2_2}
	\begin{center}
		\begin{tabular}{@{}cc|cccccccc@{}}
			\toprule
			M &	K & lm & lasso & ridge & elastic net &  $\ell_\alpha$ & glmboosting & rf &svm \\
			\midrule
 & &			\\
			& &\multicolumn{8}{c}{low correlation}\\
 & &			\\
			\multirow{3}{*}{100}&
 50 & 4.68 & 0.95 & 1.12 & 1.05 & 1 & 1.11 & 1.04 & 1.18 \\ 
	&100 & NA & 0.98 & 1.26 & 0.98 &  1 & 1.02 & 1.10 & 1.28 \\ 
	& 500 & NA & 0.98 & 1.13 & 0.98 &  1& 1.28 & 0.78 & 1.19 \\
 & &	\\
	\multirow{3}{*}{500}&
	50 & 1.11 & 0.97 & 0.99 & 0.95 &  1& 0.99 & 0.38 & 0.82 \\ 
	&100 & 1.07 & 0.95 & 0.90 & 0.92 &  1& 0.99 & 0.32 & 0.67 \\ 
	&500 & NA & 0.95 & 1.09 & 0.90 &  1 & 1.05 & 0.77 & 1.59 \\
 & &	\\
	\multirow{3}{*}{1000}&
	50 & 1.07 & 0.99 & 1.07 & 0.99 &  1 & 1.01 & 0.65 & 0.90 \\ 
	&100 & 1.08 & 1.00 & 1.01 & 0.95 &  1 & 1.01 & 0.63 & 0.95 \\ 
	&500 & 7.447 & 0.95 & 1.06 & 0.89 & 1 & 1.01 & 0.51 & 1.25 \\
 & &\\
			& &\multicolumn{8}{c}{high correlation}\\
	 & &		\\
\multirow{3}{*}{100}&
	50 & 118.3 & 1.00 & 1.14 & 1.02 & 1 & 1.18 & 1.11 & 1.23 \\
	&100 & NA & 0.94 & 1.09 & 1.01 &  1 & 1.03 & 0.48 & 1.05 \\ 
&500 & NA & 0.98 & 1.13 & 0.98 &  1 & 1.28 & 0.78 & 1.19 \\ 
 & &\\
\multirow{3}{*}{500}&
	50 & 1.07 & 1.00 & 1.05 & 1.01 &  1 & 0.99 & 1.02 & 1.18 \\ 
	&100 & 1.28 & 0.99 & 1.07 & 1.05 & 1 & 1.05 & 0.31 & 0.94 \\ 
&500 & NA & 1.00 & 1.79 & 1.03 &  1 & 1.13 & 0.34 & 2.4 \\ 
& & \\
\multirow{3}{*}{1000}&
	50 & 1.03 & 1.00 & 1.04 & 1.01 & 1 & 1.00 & 0.55 & 0.77 \\ 
	&100 & 1.26 & 1.14 & 1.24 & 1.1 &  1 & 1.08 & 1.33 & 1.47 \\ 
&500 & 41.07 & 1.00 & 1.39 & 1.04 &  1& 1.04 & 0.50 & 1.63 \\ 
			\bottomrule
		\end{tabular} 
	\end{center}
	\smallskip
	\footnotesize
	Note: In this case, we have two continuous and two binary predictors as true predictors for 100 customers in each market. We report the ratio of RMSE of other models compared to the alpha norm when applied to same datasets. We use five-fold cross-validation to choose tuning parameters, and we want to see how the alpha norm outperforms other models in different sample sizes for $M$ and $K$.
\end{table}

%%%%%%%%%%%%%%%%%%%%%%%%%%%%%%%%%%%%%%%%%%%%

\begin{table}[!ht]
	\caption{Model Comparison for Discrete-Choice Models II}\label{ratio_2_20}
	\begin{center}
		\begin{tabular}{@{}cc|cccccccc@{}}
			\toprule
		M &	K & lm & lasso & ridge & elastic net & $\ell_\alpha$ & glmboosting & rf &svm \\
			\midrule
			 & &\\
		& &\multicolumn{8}{c}{low correlation}\\
	 & &	\\
\multirow{3}{*}{100} &
	50 & 1.22 & 1.19 & 0.68 & 1.18 & 1 & 0.88 & 0.49 & 0.67 \\ 
	&100 & NA & 0.95 & 1.03 & 0.97 &1& 0.97 & 0.55 & 1.01 \\ 
	&500 & NA & 1.01 & 1.31 & 0.98 & 1 & 0.94 & 1.74 & 2.41 \\
 & &	\\
	\multirow{3}{*}{500}&
	50 & 1.20 & 1.03 & 1.19 & 1.05 &  1 & 1.05 & 0.60 & 1.12 \\
	&100 & 1.18 & 0.97 & 0.99 & 0.92 &  1 & 1.02 & 0.56 & 0.74 \\ 
	&500 & NA & 1.03 & 1.28 & 1.06 &  1& 1.27 & 0.60 & 1.31 \\ 
 & &	\\
	\multirow{3}{*}{1000}&
	50 & 1.03 & 0.99 & 1.03 & 1.00 &  1& 0.99 & 1.24 & 1.10 \\ 
	&100 & 1.15 & 1.02 & 1.12 & 1.03 &  1 & 1.03 & 0.50 & 0.99 \\ 
	&500 & 17.03 & 0.96 & 1.34 & 0.97 & 1 & 1.18 & 0.84 & 1.42 \\ 
		 & &	\\
		& &\multicolumn{8}{c}{high correlation}\\
	 & &	\\
		\multirow{3}{*}{100}&
	50 & 10.41 & 1.05 & 1.37 & 1.12 &  1 & 1.19 & 0.69 & 1.22 \\ 
	&100 & NA & 1.32 & 2.17 & 1.29 &  1& 1.24 & 2.2 & 2.08 \\ 
&500 & NA & 0.88 & 0.93 & 0.88 &  1 & 0.91 & 0.49 & 1.04 \\
 & &\\
\multirow{3}{*}{500}&
	50 & 1.19 & 1.01 & 1.12 & 1.00 & 1 & 1.00 & 0.55 & 0.93 \\ 
	&100 & 1.27 & 0.98 & 1.05 &0.98 &  1& 1.00 & 0.49 & 0.99 \\ 
	&500 & NA & 1.09 & 1.52 & 1.07 & 1 & 1.16 & 0.50 & 1.54 \\ 
 & &	\\
	\multirow{3}{*}{1000}&
	50 & 1.01 & 1.04 & 1.06 & 1.03 &  1 & 1.00 & 0.92 & 0.96 \\ 
	&100 & 1.17 & 1.00 & 1.02 & 0.99 &  1 & 1.06 & 0.43 & 0.80 \\ 
	&500 & 16.18 & 0.99 & 1.33 & 1.00 &  1 & 1.05 & 0.55 & 1.71 \\ 	
			\bottomrule
		\end{tabular} 
	\end{center}
	\smallskip
	\footnotesize
	Note: In this case, we have two continuous and 20 binary predictors as true predictors for 100 customers in each market. We report the ratio of RMSE of other models compared to the alpha norm when applied to same datasets. We use five-fold cross-validation to choose tuning parameters, and we want to see how the alpha norm outperforms other models in different sample sizes for $M$ and $K$.
\end{table}

\begin{table}[!ht]
\caption{List of Marketing Predictors}\label{meaning-variables}
\centering
\begin{tabular}{ll}
\toprule
Variable & Description\\
\midrule
$Y_{jmt}$ &  log of quantity/units of product j sold in store m at week t , i.e, \(log(q_{jmt})\) \\
\(X_{\mbox{price},jmt}\) &  log of price of products at time t in store  m \\
\(X_{\mbox{product},jmt}\) & vector of attributes of the product j in store m at time t, \\
& such as brand, volume, flavor, cut type, cooking method, package size, fat, salt levels\\
\(X_{\mbox{promotion},jmt}\) &  vector of promotional variables of  product j in store m at time t,\\
&such as promotions, display, and features \\
\(X_{\mbox{week},t}\) &  label of which week the record represents \\
\(X_{\mbox{store}}\) & reference information of store, such as equivalized volume and iri\_key\\
\bottomrule
\end{tabular}
\end{table}

\begin{table}[!ht]
\caption{Out-of-Sample RMSE Ratio and $R^2$ of Different Models }\label{market_rmse}
\begin{center}
\begin{tabular}{l|cccccccc}
\toprule
Method & lm & glmboosting & rf & svm & lasso & ridge &elastic net& $\ell_\alpha$\\
\midrule
%RMSE & 1.005 & 0.979 & 0.950 & 0.974 & 0.981 & 0.989& 0.980 & 0.979 \\
RMSE ratio & 1.027 & 1.000 & 0.970 & 0.995 & 1.002 & 1.010 & 1.001& 1\\
$R^2$ & 0.325 & 0.360& 0.398 & 0.367 & 0.358 & 0.347& 0.359 & 0.360\\
	\bottomrule
\end{tabular}
\end{center}
\smallskip
\footnotesize
Note: Here we show the out-of-sample RMSE ratio and $R^2$ of all models compared to the alpha norm when applied to $\mbox{DGP}^{(2)}$. We can see in this case that the alpha norm performs the best among the linear models. The alpha norm, lasso, and ridge use the same $\lambda$ here. The best  $\ell_\alpha$ model selects 158 predictors (intercept included) in total, and $\alpha=0.9, \lambda=1$. The RMSE of the alpha norm is 0.979 in this analysis.
\end{table}

%\begin{table}[!ht]
%\caption{Average RMSE Ratio and $R^2$ of Different Models When Estimating Log(Lift) }\label{log_Lift_est}
%\begin{center}
%\begin{tabular}{l|cccccccc}
%\toprule
%Method & lm & glmboosting & rf & svm & lasso & ridge &elastic net& $\ell_\alpha$\\
%\midrule
%RMSE ratio & 1.069 & 0.999 & 1.004 &  0.999 & 1.008 & 0.995 & 0.972 & 1\\
%$R^2$  & 0.063 & 0.073 &  0.063 &  0.072 &  0.056 &  0.080 &
% 0.122 & 0.0710 \\
%	\bottomrule
%\end{tabular}
%\end{center}
%\smallskip
%\footnotesize
%Note: Here we show the average RMSE and $R^2$ in a two-fold validation test when predicting log(lift) for all models.
%\end{table}

%%%%%%%%%%%%%%%%%%%%%%%%%%%%%%%%%%%%%%%%%
\clearpage
\newpage

\end{document}